# Optimizing Machine Translation through Prompt Engineering:
# An Investigation into ChatGPT's Customizability

**Masaru Yamada**  masaru.yamada@rikkyo.ac.jp
College/Graduate School of Intercultural Communication,
Rikkyo University, Tokyo, Japan

**Abstract**

This paper explores the influence of integrating the purpose of the translation and the target audience into prompts on the quality of translations produced by ChatGPT. Drawing on previous translation studies, industry practices, and ISO standards, the research underscores the significance of the pre-production phase in the translation process. The study reveals that the inclusion of suitable prompts in large-scale language models like ChatGPT can yield flexible translations, a feat yet to be realized by conventional Machine Translation (MT). The research scrutinizes the changes in translation quality when prompts are used to generate translations that meet specific conditions. The evaluation is conducted from a practicing translator's viewpoint, both subjectively and qualitatively, supplemented by the use of OpenAI's word embedding API for cosine similarity calculations. The findings suggest that the integration of the purpose and target audience into prompts can indeed modify the generated translations, generally enhancing the translation quality by industry standards. The study also demonstrates the practical application of the "good translation" concept, particularly in the context of marketing documents and culturally dependent idioms.

## 1. Introduction

With the advent of ChatGPT, a wider audience, including translation service providers, translators, and non-NLP engineers, can now readily experience the Large Language Model (LLM) capabilities of GPT. Furthermore, GPT-4 has proven to be considerably more powerful than its predecessor, GPT-3.5, overall. This improvement extends to translation tasks using ChatGPT.

In this context, prompt engineering, which is recognized as the key to harnessing the full potential of ChatGPT, also becomes crucial for achieving optimal performance from the LLM. Certainly, proficiency in prompt engineering is required to utilize ChatGPT effectively for translation. Despite it being only a few months since the release of ChatGPT-4 (plus), several papers exploring prompt design to enhance translation efficiency and quality have already been published. However, most of these studies are authored by Natural Language Processing (NLP) researchers, often exploring topics like how ChatGPT prompts might overcome traditional NLP challenges. Consequently, many of these studies aim to improve translation accuracy using ChatGPT prompts, distinguishing our study from theirs.

The primary objective of this research is to investigate the feasibility of generating machine translation outputs suitable for professional translators through the inclusion of suitable prompts in ChatGPT, referencing previous translation studies strategies, translation practices, and ISO documents. Specifically, we aim to examine how the translation output changes when explicitly prompted with translation strategies explored in translation research (e.g., dynamic equivalence vs. formal correspondence). Secondly, we will prompt ChatGPT with a translation specification encompassing parameters such as the purpose of the translation and the target audience, to facilitate more flexible integration of generated translations into professional translation workflows. We will examine how these prompts impact the translation process.

Let us illustrate with an example. The Japanese expression, *We are friends who ate out of the same pot*, is currently translated by most MTs verbatim. This literal translation may be suitable if the goal is to help an audience interested in Japanese culture understand unique Japanese expressions. However, adhering to recent translation norms, the phrase might be translated as *We have been through thick and thin together*. In other words, translation isn't a random process; it's determined by the purpose, the target audience, and the established translation strategy. The translation is performed based on this strategy. In this research, we introduce this step-by-step process into ChatGPT via prompts and assess the resulting translation changes.

## 2. Literature Review

Jiao et al. (2023) analyze ChatGPT's machine translation (MT) competence, revealing it competes with commercial systems in high-resource European languages but struggles with low-resource or distant ones. Notably, pivot prompting enhances translation quality significantly, and ChatGPT fares well in spoken language translation. GPT-4's incorporation further bolsters its capabilities. Gao et al. (2023) develop a new method for translation prompts, involving translation task information, context domain information, and Part-of-Speech tags, improving ChatGPT's performance. Their approach outperforms commercial systems in multiple translation directions using few-shot prompts.

Moslem et al. (2023) underscore the value of consistency and domain-specific terminology adaptation, demonstrating how large-scale language models can improve real-time adaptive MT, particularly for high-resource languages, while aiding less-supported languages via a combination of strong encoder-decoder models and fuzzy matches. Gu (2023) tackles the issue of translating Japanese attributive clauses to Chinese, a consistent problem in current tools. They introduce a pre-edit scheme and a two-step prompt strategy using ChatGPT, achieving significant improvement in average translation accuracy.

Despite these developments, the focus remains primarily on language-specific, low-resource language pairs, and problems conventional MT can't resolve. Industrial translation and pragmatic considerations, such as translation practice and ISO guidelines, aren't well-studied. This highlights a need for future research to address these practical aspects.

## 3. Translation Pre-Production Process

The purpose of this study, as stated in the introduction, is to enhance the quality of MT through a highly customizable approach using ChatGPT and prompts, which can seamlessly integrate into the Language Service Provider/Translation Service Provider (LSP/TPS) workflow of a practical translation job and also serve as a better computer-assisted translation (CAT) tool for the professional translators. To achieve this, it is crucial to understand the fundamental structure of the translation processes in the industry. Given the limited space, this paper will focus on extracting and elucidating the core concepts.

The translation production process, defined by ISO 17100 (ISO, 2012), is subdivided into three primary stages: pre-production, production, and post-production. The pre-production stage is particularly significant here. It involves defining and determining the requirements of the final translation before the actual translation production process begins. In practical terms, translation does not start immediately after the client provides the source text. Instead, at this stage, the translation project manager, among others, decide on the specifications of the translation, which includes the purpose of the translation, the intended audience, whether there will be a designated glossary, whether the translation will adhere to a style guide and other such aspects. Various documents detail these considerations (ASTM, 2014, ISO 2012, 2017, Onishi and Yamada, 2021). Some of these documents outline the per-word cost and whether a CAT tool should be used. However, these are not items for inclusion in the ChatGPT prompts. Relevant items for the prompts are:

Items related to the target language:

- What is the purpose of the translation?
- Who is the intended audience for the target sentence?
- What is the locale of the target language?
- What is the register?
- Is a style guide available?

The idea that the quality of translation required fluctuates as the translation specifications change is well-established in Multidimensional Quality Metrics (MQM)[1] and Dynamic Quality Framework (DQF)[2]. This notion is widely accepted among TSPs and working translators. However, it has not been a primary concern for the NLP researchers developing MT. Thus far, MTs have not incorporated functions to set or adjust the specifications of the translation as described above, making it challenging for current MT to fit into practical translation workflows.

In response, the Machine Translation Post-Editing (MTPE) method is often adopted in practical translation to modify MT translations. In light of the above discussion, post-editing can be seen as a process to fill in the gaps in translation output due to the MT's failure to set the previously mentioned requirements. In essence, the MT translates the given source text without comprehending the translation specifications, including the purpose of the translation. Consequently, human post-editors are tasked with correcting the MT translation to bridge these gaps.

## 4. The objective of this paper and method of evaluation

### 4.1. Objective

The primary objective of this paper is to evaluate the transformation in the translation output when prompts, typically determined during the pre-production phase of translation as per practical translation norms and translation studies outlined in ISO and other standards, are supplied to ChatGPT. The generated translations will be subject to qualitative analysis. Furthermore, the resemblance of the translations to the source text will be examined using the cosine similarity metric applied to the embeddings obtained from OpenAI's embedding API[3]. Note that this validation only deals with a very limited number of samples. The reason for this, however, is that it does not rely on automatic evaluation methods such as the BLEU score.

### 4.2. Cosine Similarity

Cosine similarity is a method utilized to measure the similarity between two vectors within a high-dimensional space. Essentially, it quantifies how closely two vectors point in the same direction by computing the cosine of the angle between them. For the purposes of this paper, we utilized the *text-embedding-ada-002* provided by OpenAI to calculate the cosine similarity between the source and target texts. This API employs a GPT-3-based model capable of understanding text content and generating representative vectors. These vectors are able to encapsulate semantic similarity, indicating whether the texts express comparable content[4].

---

[1] https://themqm.org/
[2] https://www.taus.net/resources/blog/category/dynamic-quality-framework
[3] https://platform.openai.com/docs/guides/embeddings

**4.3. Validity and Reliability of Using this Method for Translation Evaluation**

Theoretically, it is feasible to gauge the semantic similarity between source and target texts by calculating the cosine similarity of vectors generated using *ada-002*. This technique could be particularly beneficial in determining the accuracy with which a translation preserves the meaning of the source text. However, there are several significant considerations.

     First, the quality of a translation should not solely rest on semantic accuracy but should also factor in cultural subtleties and language-specific elements, such as idioms. Moreover, a direct translation is not always the optimal choice. A translation that strictly retains the original meaning may not necessarily constitute a good translation. Different languages have distinct grammatical structures and expressive modes; hence, translations must fully leverage the characteristics of each language. Therefore, this paper also includes a qualitative evaluation performed by a human expert, the author, in conjunction with the aforementioned method.

**5. Verification Prompts**

Two prompts were utilized for verification purposes. The first prompt is detailed below. All parameters, barring two—Purpose of the translation and Target readers of the translation—were left unset. This was grounded in the author's experience as a professional translator, leading to the conclusion that these two parameters are essential even in everyday translation work. Other items, like a predetermined glossary, are often not provided by the client.

| |
|---|
| Translate the following Japanese [source text] into English. Please fulfill the following conditions when translating. |
| Purpose of the translation: *You need to fill in.* |
| Target audience: *You need to fill in.* |
| [source text] *You need to fill in.* |

     Table1: Prompts Specifying the Purpose of Translation and Target Audience

**6. Comparative Verification Method**

Three reference translations were prepared for comparison: DeepL, Google Translate, and ChatGPT Plus (GPT4) simply using the prompt "Translate to English." The ChatGPT translations, further enhanced by adding the prompts "purpose" and "target reader," were compared to these. We also added the prompt *Please generate three translations* to produce three distinct translations for comparison. For translators in practice, the ability to consider multiple potential translations is crucial.

     Pym (1992) posits that the expertise of a skilled translator lies in their capacity to generate as many possible translations of a source text and select the best one— one that adheres to the translation specifications (in this case, the purpose of translation and the target audience). In other words, it is the skill to select the translation that most aptly fulfills the translation specifications. From this perspective, prompting ChatGPT to generate multiple translations aligns well with practical translation methods.

---

[4] Script to calculate the consine similarity is available at
https://github.com/chuckmy/chatGPT_Cosine-Similarity

## 7. Verification Result 1

The initial step in the verification process involved translating a fictitious marketing statement from Japanese into English. The purpose of the translation and the target audience are outlined below.

---

Translate the following Japanese [source text] into English. Please fulfill the following conditions when translating.

Purpose of the translation: *To market our own brand of cosmetics and to be displayed on our website*
Target audience: *Women in their 20s*

[source text] 私たちが開発したファンデーションはあなたの自然な美しさを引き立てます。シームレスに肌に溶け込み、まるで素肌そのもののような仕上がりを提供します。

---

Table 2: Actual promt specified.

The given statement is part of an advertisement for a women's cosmetic foundation and is typically classified as a "marketing" text type. Translations for marketing differ substantially from those for, say, instruction manuals or contracts. Marketing translations necessitate an understanding of the cultural nuances and conventions of the target market and the crafting of an enticing message. This involves creativity and maintaining brand consistency. Conversely, translations of instruction manuals and contracts must be precise and direct, emphasizing the accurate conveyance of technical details and legal provisions. Therefore, marketing translations demand cultural adaptability, persuasiveness, creativity, and brand consistency, while translations of instruction manuals and contracts prioritize clarity and information accuracy.

Keeping this in mind, it is appropriate to designate the translation purpose as above: *To market our own brand of cosmetics and to be displayed on our website*. Similarly, it is self-evident that the target audience for this translation is *women in their 20s*, which aligns with the product's target demographic. In professional translation practice, the information about the purpose and target audience, as indicated here, will be communicated to the translator. In fact, translating without such contextual information would be deemed unprofessional.

Table 3 presents the three types of translations obtained by employing the above-mentioned prompts. The top three in the table—DeepL (DL), Google Translate (GT), and ChatGPT Plus (GPT) with translation instructions only—are the reference translations for this analysis. The v1, v2, and v3 indicate the three translations generated by the prompt above. 'C.S.' represents 'cosine similarity,' and 'Rank' signifies the similarity order from the most similar to the original source test. The closest to the source is ranked 1 (first).

Given that this translation is for a cosmetic product's marketing, the English translation should also be "stylish". The final portion of DL and GT's translations, *skin itself*, is rather literal. A more unique English translation for this part of the sentence would be preferable. In this aspect, GPT's translation is generally more natural, and its use of the phrase *looks like ... bare skin* at the end outperforms existing MT engines.

Reviewing the types of translations generated by the above prompt, all three translations differ from the baseline translation, an effect attributable to the prompts. However, both v1 and v2 resemble the baseline translation as they employ an expression similar to *enhance ... natural beauty* in the first sentence. In the latter part of the sentence, v1 and v2 also contain *bare skin*, aligning them with the baseline translation. Conversely, the v3 sentence employs a different phrase, *serves to amplify your inherent beauty* instead of *enhance*, and uses *natural skin* in the latter part, distinguishing it from the existing translations. From this standpoint, a professional translator is more likely to select v3 as the fit-for-purpose translation.

Interestingly, the cosine similarity of these sentences revealed that the v3 translation was closest to the source text. The second highest was the ChatGPT baseline translation, while the DL and GT translations exhibited lower values than v1 and v3. This suggests that the appropriate translation choices from a professional translator's standpoint generally align with the cosine similarity ranking. The translation prompted by this marketing translation, considering its purpose and target audience, is likely to be chosen as a more suitable translation than one from an existing translation engine. In other words, this implies that ChatGPT, when combined with suitable prompts, might be able to offer translations that adhere to the specifications outlined in the translation production process.

[source text]
私たちが開発したファンデーションはあなたの自然な美しさを引き立てます。シームレスに肌に溶け込み、まるで素肌そのもののような仕上がりを提供します。

| Type | Target sentence | C.S. | Rank |
|---|---|---|---|
| DL | Our foundations enhance your natural beauty. They blend seamlessly into the skin and provide a finish that looks like your skin itself. | 0.861 | 5 |
| GT | Our foundations are designed to enhance your natural beauty. It blends seamlessly into the skin and provides a finish that looks like bare skin itself. | 0.868 | 3 |
| GPT | The foundation we developed enhances your natural beauty. It seamlessly blends into your skin, providing a finish that feels just like your own bare skin. | 0.873 | 2 |
| v1 | Our newly developed foundation enhances your natural beauty. It blends seamlessly into your skin, providing a finish that's just like your own bare skin., | 0.870 | 3 |
| v2 | Experience the natural beauty enhancement with our specially designed foundation. Its unique formulation blends effortlessly into your skin, giving the impression of flawless, bare skin., | 0.863 | 4 |
| v3 | The foundation we've created serves to amplify your inherent beauty. Seamlessly melting into your skin, it leaves you with a finish indistinguishable from your natural skin. | 0.875 | 1 |

Table 3: Results of translations generated with the prompt.

## 8. Verification 2

Consider another dimension of translation involving culture-dependent idioms and expressions, as initially discussed in this paper. The original source phrase, *we are friends who ate rice from the same pot*, is a Japanese idiom symbolizing a relationship between individuals who have weathered difficult situations together or shared profound experiences. While the literal meaning involves sharing rice cooked in the same pot, the figurative implication refers to enduring hardships together or strengthening bonds through shared experiences.

When translating, such expressions are often embedded in the original text. However, as a direct translation may not be comprehensible, the translation objective is typically set to render these expressions naturally intelligible to English speakers unfamiliar with Japanese culture.

To meet this translation objective, professional translators working in the field would adapt the expression. However, existing MT systems, which do not align with this goal, often produce literal translations that are unfit for use. Therefore, to achieve this objective in a practical working scenario, we set the following prompts:

> Translate the following Japanese [source text] into English. Please fulfill the following conditions when translating.
> Purpose of the translation: *Use natural expressions that can be understood by English speakers who are not very familiar with Japanese culture.*

> Target audience: *General English-speaking audience.*
> [source text] 私たちは同じ釜の飯を食べた仲です。

<div align="center">Table 4: Actual prompt specified for cultural-bound expression.</div>

Comparing the three translations acquired with the baseline translations: all translations for DeepL (DL), Google Translate (GT), and ChatGPT Plus (GPT) are essentially literal translations of the source text. In contrast, the translations from v1 through v3, which were generated with prompts, exhibit differences. Nevertheless, the v1 translation still includes *rice* and *the same pot*, differing little from the baseline translation. Consequently, from a professional translator's perspective, v1 does not represent a superior translation in this context. In contrast, v2 and v3 are translations that English speakers could comprehend. Specifically, v2, with *been through thick and thin together*, is deemed appropriate in terms of target acceptability.

Regarding cosine similarity, v2 secures the second position, affirming its semantic proximity. Hence, the result reasserts that including appropriate prompts for purpose and target reader yields a more suitable translation.

[source text] 私たちは同じ釜の飯を食べた仲です。

| Type | Target sentence | C.S. | Rank |
|---|---|---|---|
| DL | We are friends who ate out of the same pot. | 0.772 | 1 |
| GT | We ate rice from the same pot. | 0.727 | 5 |
| GPT | We ate rice from the same pot. | 0.727 | 5 |
| v1 | We have shared the same pot of rice. | 0.743 | 4 |
| v2 | We have been through thick and thin together. | 0.759 | 2 |
| v3 | We've broken bread together. | 0.744 | 3 |

<div align="center">Table 5: Results of translations of cultural-bound expression</div>

## 9. Prompt for Dynamic Equivalence

Dynamic equivalence (Nida, 1969/2003) is a translation strategy aiming to balance the reader's response between the source and target texts. Take, for example, the term *Lamb* in the source text, which would be directly translated as *lamb*. However, if the target audience is in Iceland, where sheep are not indigenous, translating the word as *lamb* may not convey the intended nuance. Therefore, to equate the original text reader's reaction with that of the translated text reader, a strategy is employed to transfer the meaning (nuance) of the original text by substituting *lamb* with *seal*. This is somewhat akin to replacing *ate (rice) from the same pot* with *have been through thick and thin together* from the previous example, albeit more extreme.

We aim to examine whether supplying dynamic equivalence prompts to ChatGPT can yield more flexible and creative translations. Presented below are the actual prompts and the source text.

> Dynamic equivalence is a strategy for translating from the perspective of equalizing the reader's response to the [source text] and the [target text]. In the example below, the word "Lamb" in the original text would be translated as "lamb" in the literal translation. However, when translating for Iceland, which has no sheep, it is difficult to convey the nuance of the word "lamb". From the standpoint of equalizing the reader's reaction, this is a ruse to translate it as "seal". It is believed that "lamb" in the [source text] and "seal" in the [target text] will evoke the same reaction in the reader.
> [source text] Lamb of God
> [target text] Seal of God

> Following this concept and example, please translate the following [source text] into English using the dynamic equivalent. Please replace the translation with something that would be understood in an English-speaking culture.
> [source text] 彼女の歌声は美空ひばりを彷彿とさせる。

Table6: Actual prompt for dynamic equivalence

The source text translates to *Her singing voice reminds me of Hibari Misora*. The focal point here is the singer H*ibari Misora*. Without understanding who *Hibari Misora* is, the original sentence's meaning becomes unclear. Notably, Hibari Misora (May 29, 1937 - June 24, 1989) was a prominent Japanese singer and actress, revered for her powerful voice and emotional performances. She is widely recognized as a definitive figure in the traditional Japanese "enka" country music style.

After defining the dynamic equivalence, the prompt states, *please replace the translation with something that would be understood in an English-speaking culture*. Who could be a suitable replacement?

[source text] 彼女の歌声は美空ひばりを彷彿とさせる。

| Type | Target sentence | C.S. | Rank |
|---|---|---|---|
| DL | Her singing voice is reminiscent of Hibari Misora. | 0.876 | 1 |
| GT | Her singing voice is reminiscent of Hibari Misora. | 0.876 | 1 |
| GPT | Her singing voice reminds me of Misora Hibari. | 0.873 | 2 |
| v1 | Her singing voice evokes memories of Judy Garland. | 0.830 | 3 |
| v2 | Her singing voice is reminiscent of Billie Holiday. | 0.823 | 5 |
| v3 | Listening to her sing, one can't help but think of Ella Fitzgerald. | 0.826 | 4 |

Table 7: Results of dynamic equivalence prompt

The table reveals that the baseline translations from DeepL (DL) and Google Translate (GT) are identical, and the GPT translations are nearly the same. The part mentioning *Hibari Misora* is directly translated. However, none of the baseline translations meet the dynamic equivalence strategy's objectives. On the other hand, versions v1 through v3 substitute *Hibari Misora* with *Judy Garland*, *Billie Holiday*, and *Ella Fitzgerald*, respectively. These are all interesting choices. In this instance, all translations opted for famed American singers. While the translations succeed in fulfilling the dynamic equivalence goal, the selection of these particular singers, who share a similar era with the original singer, is indeed apt. Yet, from a translator's perspective, it's debatable which—v1 *Judy Garland*, v2 *Billie Holiday*, or v3 *Ella Fitzgerald*— achieves the dynamic equivalence of the original singers most effectively.

Reviewing the cosine similarity rankings, the baseline translations simply transliterate *Hibari Misora* as it is ranked 1st and 2nd. These translations score high on cosine similarity as they semantically align with the source text. However, they do not meet this study's objective. On the other hand, v1, v3, and v2 rank 3rd, 4th, and 5th, respectively. According to this order, v1, which substitutes *Hibari Misora* with *Judy Garland*, achieves the highest score (3rd) among the three alternatives. However, this doesn't necessarily mean that Judy Garland is the singer most similar to *Hibari Misora* based on cosine similarity; this score considers the entire sentence.

Therefore, to determine which of the three English-speaking singers most closely resembles *Hibari Misora* in terms of cosine similarity, we standardized the translation sentence structure to *Her singing voice is reminiscent of ...* as generated by DL/GT. We then replaced only the singer's name and analyzed the cosine similarity. The results are outlined in the table below.

Naturally, the baseline translation takes the top spot. However, v3's *Ella Fitzgerald* ranks second, suggesting she is most similar to *Hibari Misora* in terms of cosine similarity. Of course, this doesn't guarantee that the most appropriate dynamic equivalent is *Ella Fitzgerald*. Yet, it does hint at a potential overlap with the rationale employed by professional translators in selecting suitable translations in practical translation. Conversely, these validation results also suggest that it may be advantageous for professional translators to view multiple translation candidates and cosine similarities as part of their Computer-Assisted Translation (CAT) tool.

| Type | Target sentence | C.S. | Rank |
|---|---|---|---|
| DL | Her singing voice is reminiscent of Hibari Misora. | 0.876 | 1 |
| v1 | Her singing voice is reminiscent of Judy Garland. | 0.826 | 3 |
| v2 | Her singing voice is reminiscent of Billie Holiday. | 0.823 | 4 |
| v3 | Her singing voice is reminiscent of Ella Fitzgerald. | 0.833 | 2 |

Table 8: To find out the best 'equivalence' to *Hibari Misora*

## 10. Conclusion

In this study, we investigated how incorporating the purpose of the translation and the target audience into prompts - key elements of translation specifications defined in the front-end phase of the industrial translation production process - alters the quality of translations generated by ChatGPT. We examined shifts in translation quality when prompts were issued to generate translations meeting these specific conditions. The assessment was conducted from the perspective of a practicing translator, both subjectively and qualitatively. Additionally, we utilized Open AI's word embedding API to calculate cosine similarity, complementing our qualitative evaluation.

Our findings indicated that incorporating the purpose and target readers into prompts indeed altered the generated translations. This transformation, aimed at satisfying the criteria of purpose and target audience, generally improved the translation quality by industry standards. Particularly for marketing documents and culturally dependent idioms, a translation strictly faithful to the source text is not regarded as a "good translation." Thus, by appropriately setting the purpose of the translation, the prompt can be adjusted to produce a translation that is closer to the target culture, favoring intent translation and domestication. This demonstrates the practical application of the "good translation" concept.

Furthermore, we employed strategies like dynamic equivalence, replacing items with high cultural dependency on the target audience. Specifically, we substituted the renowned Japanese singer *Hibari Misora* with *Ella Fitzgerald*, a singer considered her equivalent in English-speaking countries, facilitating a high degree of creative translation. Our results confirmed that prompts can progressively guide creative translations, such as replacing *Hibari Misora* with *Ella Fitzgerald*. These translations may prove more useful than conventional machine-translated ones, even serving professional translators as preliminary translations for post-editing.

Although our study was limited by the development of prompts and a small sample size, further, larger-scale experiments and validations are required to practically apply the results. Nevertheless, our findings confirmed that incorporating appropriate prompts into large language models like ChatGPT can generate flexible translations, an accomplishment yet to be achieved by traditional MT. We anticipate future verifications with great interest.